\documentclass[conference]{IEEEtran}
\IEEEoverridecommandlockouts
\usepackage{cite}
\usepackage{amsmath,amssymb,amsfonts}
\usepackage{algorithmic}
\usepackage{graphicx}
\usepackage{textcomp}
\usepackage[table,xcdraw]{xcolor}
\def\BibTeX{{\rm B\kern-.05em{\sc i\kern-.025em b}\kern-.08em
    T\kern-.1667em\lower.7ex\hbox{E}\kern-.125emX}}
\begin{document}

\title{Affect Models Have Weak Generalizability to Atypical Speech\\
}

\author{\IEEEauthorblockN{Jaya Narain} 
\IEEEauthorblockA{\textit{Apple} \\
Cupertino, CA \\
jnarain@apple.com}
\and
\IEEEauthorblockN{Amrit Romana} 
\IEEEauthorblockA{\textit{Apple} \\
Cupertino, CA \\
ak\_romana@apple.com}
\and
\IEEEauthorblockN{Vikramjit Mitra} 
\IEEEauthorblockA{\textit{Apple} \\
Cupertino, CA \\
vmitra@apple.com}
\and
\IEEEauthorblockN{Colin Lea} 
\IEEEauthorblockA{\textit{Apple} \\
Cupertino, CA \\
colin\_lea@apple.com}
\and
\IEEEauthorblockN{Shirley Ren} 
\IEEEauthorblockA{\textit{Apple} \\
Cupertino, CA \\
shirleyr@apple.com}
}

\maketitle
\IEEEpeerreviewmaketitle
\begin{abstract}.  
Speech and voice conditions can alter the acoustic properties of speech, which could impact the performance of paralinguistic models for affect for people with atypical speech.  We evaluate publicly available models for recognizing categorical and dimensional affect from speech on a dataset of atypical speech, comparing results to datasets of typical speech.  We investigate three dimensions of speech atypicality: intelligibility, which is related to pronounciation; monopitch, which is related to prosody, and harshness, which is related to voice quality.   We look at (1) distributional trends of categorical affect predictions within the dataset, (2) distributional comparisons of categorical affect predictions to similar datasets of typical speech, and (3) correlation strengths between text and speech predictions for spontaneous speech for valence and arousal.  We find that the output of affect models is significantly impacted by the presence and degree of speech atypicalities.  For instance, the percentage of speech predicted as sad is significantly higher for all types and grades of atypical speech when compared to similar typical speech datasets.  In a preliminary investigation on improving robustness for atypical speech, we find that fine-tuning models on pseudo-labeled atypical speech data improves performance on atypical speech without impacting performance on typical speech.  Our results emphasize the need for broader training and evaluation datasets for speech emotion models, and for modeling approaches that are robust to voice and speech differences.
\end{abstract}


\section{Introduction and Related Work}
Systems that help interpret, track, and modulate affect have the potential to enhance day-to-day wellbeing, augment social experiences and learning, and improve the quality of interactions with devices and technology.  For instance, prior work has explored models and systems for detecting and monitoring depression, stress, and anxiety \cite{reddy2024audioinsight, cummins2023multilingual, sano2018identifying, niu2025depression, ringeval2019avec, cummins2015review}, as well as tools for improving psychological wellbeing \cite{jeong2023deploying}, improving public speaking skills \cite{saufnay2024improvement}, enhancing meeting effectiveness,  \cite{samrose2021meetingcoach}, summarizing notifications \cite{an2024emowear}, better understanding child learning with robotic agents \cite{chen2020impact, kim2024child}, and reducing stress when driving \cite{chung2019methods}.  Speech is often used as an interaction modality and data stream in such systems.  For instance, a study on digital phenotyping in real-world settings examined multilingual speech markers of depression, \cite{cummins2023multilingual}, the meeting coaching system \cite{samrose2021meetingcoach} included tracking tone and suggest tone modifications, and the notification summary tool \cite{an2024emowear} used speech as an input to recommend emotional summaries for smartwatch notifications.

State-of-the-art emotion recognition models \cite{wagner2023dawn, goncalves2024odyssey, mitra2024investigating} 
include fine-tuning general purpose audio embedding models like HuBERT \cite{hsu2021hubert, baevski2020wav2vec} and wav2vec 2.0 on affect datasets like MSP-Podcast \cite{lotfian2017building}, IEMOCAP \cite{busso2008iemocap}, and RAVDESS \cite{livingstone2018ryerson}.  These datasets are generally acted speech or speech from a particular domains (e.g., podcasts) that have been labeled by annotators.  Recent work has also explored using generative models for affect recognition using zero- and few-shot prompting \cite{latif2023can, niu2024text, amin2024wide}. 

As the use of affect-based models increases, it is critical to understand technological limitations and ensure that deployed systems are inclusive.  Prior work has characterized model fairness and robustness, particularly with respect to language, gender, and acoustic properties \cite{derington2025testing}.  For instance, an analysis of pre-trained audio models fine-tuned on MSP-Podcast looked at model fairness with respect to accent, language, pitch, linguistic sentiment, and gender and found that fairness varied between models but that most models had similar sub-group prediction distributions relative to the overall prediction distributions \cite{derington2025testing}.  A different investigation of multi-modal speech emotion fairness found gender-bias in speech emotion models - with the least bias in text-only models which contained less demographic information \cite{schmitz2022bias}.  A prior study on speaker personalization \cite{triantafyllopoulos2024enrolment} found improvements in speaker-level fairness via enrollment-based personalization. 

There has been limited prior work on the impact of atypical speech on the performance of affect models.  As part of a wider analysis of bias and speech emotion recognition, a previous study found that positive valence was more likely to be associated with speakers who did not have disabilities, and also identified potential biases related to age, gender, and other attributes \cite{slaughter2023pre}.  The study looked only at binarized valence results with speech from sixteen dysarthric speakers.  Here we examine performance differences via within- and between- dataset comparisons for different types of atypical speech with both categorical and dimensional models, using thousands of labeled speech samples from hundreds of unique speakers.

Atypical speech can be caused by conditions directly related to voice and speech including structural (e.g., voice nodules) and neurogenic conditions (e.g., spasmodic dysphonia), as well as by other long- and short-term health conditions (e.g., neuromotor conditions like ALS, and Parkinson’s; transient illnesses like colds and sore throats).   Day-to-day voice fatigue and overuse also impacts speech, causing both short- and long-term changes. Changes in voice - including harshness, breathiness, and pitch - can also occur as part of aging \cite{uclaaging}.  It is estimated that about 10\% of American adults are impacted by a voice, speech, or language condition yearly \cite{morris2016prevalence}, with an even larger impact on children \cite{nih}.  Understanding the performance of paralinguistic models on atypical speech is important to ensure systems are safe, fair, and enjoyable for the millions of speakers with speech and voice atypicalities. 

Here, we investigate the following research questions:

\textbf{RQ1:} Is the performance of speech affect models for dimensional and categorical emotions impacted by the degree and presence of speech atypicalities?

\textbf{RQ2:} Is the performance of affect models affected by specific characteristics of speech -- particularly the type of speech atypicality (harshness/atypical voice quality, intelligibility/atypical pronounciation, and monopitch/atypical prosody) and the type of speech sample (spontaneous speech, read sentences, and digital voice commands)?

\textbf{RQ3: } How is model robustness for atypical speech impacted by fine-tuning on pseudo-labeled atypical speech data and personalization via prompting?  

\section{Methods}

\subsection{Data}

Atypical speech data from Speech Accessibility Project (SAP) \cite{hasegawa2024community} dataset was analyzed.  The SAP dataset is a publicly available atypical speech dataset.  We used a subset of the data that had annotations from speech-language pathologists on types and degrees of rated atypicality.  The analyzed data included 11,184 samples from 434 speakers (284 with Parkinson's Disease, 78 with Cerebral Palsy, 53 with ALS, 16 with Down Syndrome, 2 with Ataxic Dysarthria, 1 with Flaccid Dyarthria) from three speech categories (examples in \cite{hasegawa2024community}): digital voice commands (\textit{n}=3,797; avg. length 5.00s, std. 4.65s), read novel sentences (\textit{n}=3,838; avg. length 9.03s, std. 4.42s); and  spontaneous speech samples (\textit{n}=3,549; avg. length 25.94s, std. 21.47s).   The SAP dataset includes pre-defined data splits.  We included all of the annotated SAP data in evaluation of off-the-shelf models in order to have larger sample sizes for analyses.  

\begin{figure}[b!]
  \centering
  \includegraphics[width=\linewidth]{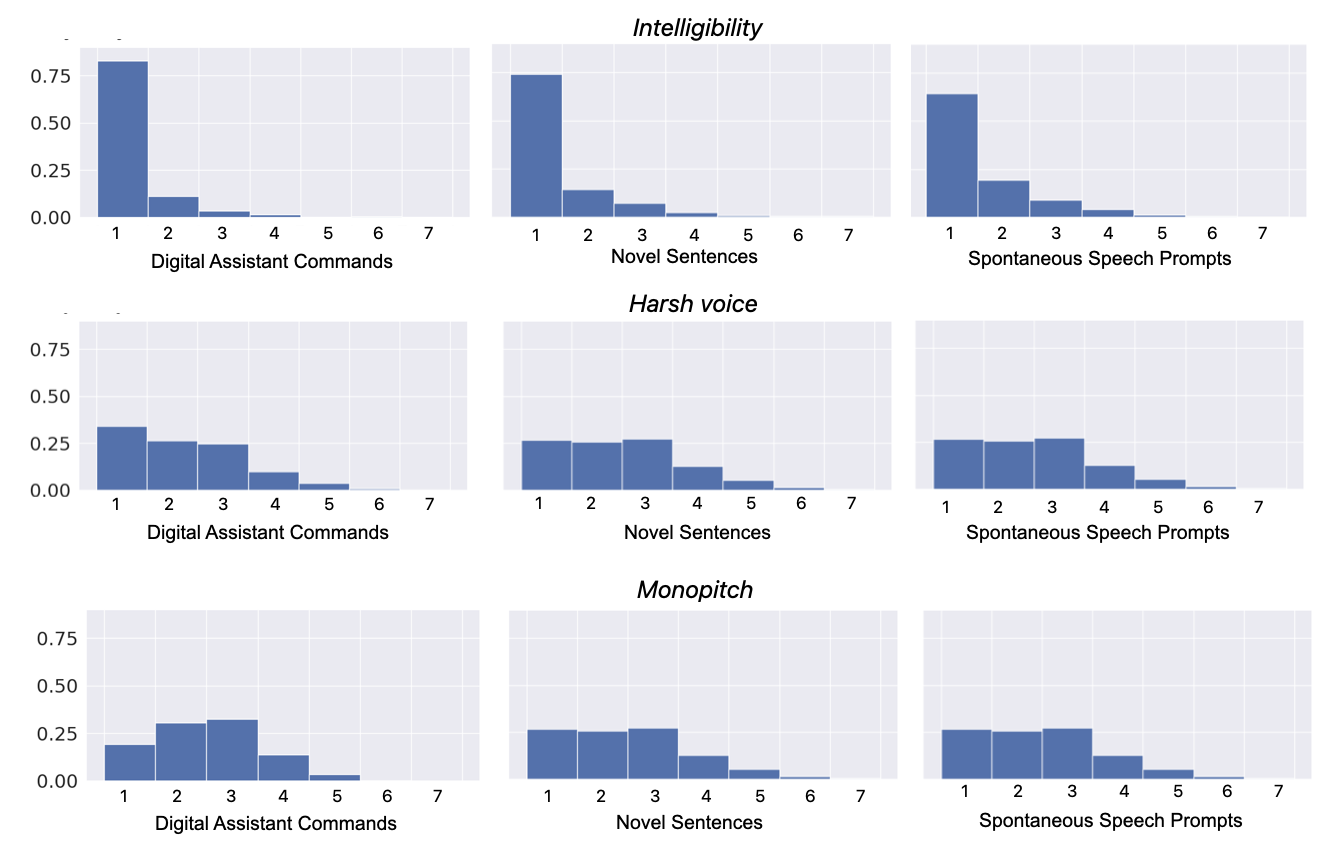}
  \caption{Distribution of annotations in dataset for each speech category (digital assistant commands, novel sentences, and spontaneous speech)}
  \label{fig:annotation_distributions}
\end{figure}

We analyzed the impact of three types of atypical speech variations on affect modeling: intelligibility, harshness, and monopitch.  These dimensions were chosen because of the availability of these ratings within the dataset and because they span a variety of speech characteristics -- intelligibility is related to pronunciation; harshness is related to voice quality; and monopitch is related to prosody.  Each analyzed sample had a rating from 1-7 for each dimension, indicating the degree of atypicality for that dimension.  For example, a rating of 7 indicates highly atypical speech with respect to intelligibility -- that is, speech with low  intelligibility.  A rating of 7 for harshness and monopitch indicates very high harshness and monopitch, respectively.  Figure \ref{fig:annotation_distributions} shows the distribution of annotations by rated speech category for intelligibility, harshness, and monopitch.

We observed some affective variation in the spontaneous speech data, where participants responded to questions (e.g., favorite books and movies) -- but, given the collection protocol, we do not expect significant distributional differences in affective variation among differing levels or types of atypicality within the dataset.  Given the elicitation prompts and from our observations listening to samples, the read speech categories (digital voice commands and novel sentences) were overwhelmingly affectively neutral.

For each speech category, we selected a dataset of typical speech with similar content for comparison: Switchboard for spontaneous speech \cite{godfrey1992switchboard}, Common Voice for read sentences \cite{ardila2019common}, and voice command audios for virtual assistants dataset (VCVA) \cite{voicecommands} for digital assistant commands.  The reference dataset for spontaneous speech may have limited relevance for experiments conducted without labels because there may be different true affective distributions between SAP and Switchboard -- in these experiments, within-dataset comparisons in SAP between levels of atypical speech are particularly informative.  The typical speech datasets for read speech (sentences and digital commands) provide strong comparison points, as the samples should generally be affectively neutral across datasets. 

The SAP dataset includes only English speech and is predominately from speakers with American accents.  We sub-sampled data from each typical speech dataset for these characteristics using available metadata on language and accent to match the domain of the SAP data as closely as possible, and also considered utterance length and word count.  The sub-sampled datasets included \textit{n}=2897 samples from the Switchboard dataset (avg. length 8.13, std. 4.02s), \textit{n}=3500 samples from the Common Voice dataset (avg. length 4.65s, std. 1.21s), and \textit{n}=3700 samples from the VCVA dataset avg. length 2.41, std. 0.58s). Note that after sampling, there are still some distributional differences among datasets.   We used the wav2vec2 forced alignment model \cite{wav2vecfa} with annotated transcriptions to trim leading and trailing silence prior to extracting embeddings (validated via spot checking a subset of samples).

\subsection{Model evaluation}

\subsubsection{Categorical Emotions}
We conducted an analyses of distributional outputs of categorical emotions, with three publicly available pre-trained categorical emotion models, and GPT-4o-audio-preview: 
\begin{itemize}
    \item \textbf{Odyssey categorical emotion model }\cite{goncalves2024odyssey}, a pre-trained WavLM model fine-tuned on MSP-Podcast data
    \item \textbf{Emotion2vec} \cite{ma2023emotion2vec}, an emotion model trained with self-supervised learning on unlabeled emotion datasets (including RAVDESS, MSP-Podcast, and other datasets) with linear layers trained on IEMOCAP
        \item \textbf{SpeechBrain} \cite{speechbrain}, a fine-tuned wav2vec2 model trained on IEMOCAP
    \item \textbf{GPT-4o-audio-preview} \cite{gpt4oaudio}, used as a zero-shot emotion rater for speech files 
\end{itemize}

We prompted GPT-4o-audio-preview to annotate categorical emotion in each sample using the following prompt, adapted from prior work\cite{niu2024text, niu2024rethinking}:
\\
\\
\textit{
``You are an emotionally-intelligent and empathetic agent. 
You will be given an audio segment with speech, and your task is to identify the primary emotion the speaker is expressing. If there is no emotion, then the primary emotion is neutral. 
Classify the speech into one of the following categories: anger, disgust, sadness, fear, surprise, happiness, neutral.
Respond with only one category and keep your responses to the category name as written and nothing else."}
\\

We assessed the percentage of each categorical emotion for each rating value and each rating type for each speech category within the SAP dataset.  We also binarized each SAP label using the rating closest the 20th percentile in the dataset for each rating type (2 for intelligibility, 4 for harsh voice and monopitch) and tabulated the percentage of categorical emotions in the binarized groupings.  We used binarized groupings for numerical comparisons to ensure a sufficient sample size in each data subset - after binarization, the analysis included 1,358 samples with low intelligibility and 2,327 with high intelligibility; 915 samples with high harshness and 2,770 with low; and 1,186 with high monopitch and 2,499 with low.  Confidence intervals are reported based on the sample size and formula for confidence intervals for proportions.  


\subsubsection{Dimensional Emotions}
For dimensional emotion modeling, we looked at correlations between dimensional emotion scores predicted from text and dimensional emotion scores predicted from audio for spontaneous speech data, which contained some natural affective variation.  We used GPT-4o to provide `pseudo-labels' for arousal and valence for each sample from text transcriptions of each file.   From labeling experiments with text-only transcriptions, we found that there was not enough variation in lexical categorical emotional content to analyze categorical correlations and so we focused the correlational analysis on dimensional emotion.

We used the Odyssey valence and Odyssey arousal dimensional models \cite{goncalves2024odyssey} (pre-trained WavLM models fine-tuned on MSP-Podcast data) to predict valence and arousal scores for all spontaneous speech data in SAP and for the Switchboard data. We did not use the read speech categories (read sentences and digital assistant commands) as there was little to no expected affective content in those categories.  We prompted GPT-4o to rate the valence and arousal of each sample using few-shot prompting with examples from the Emobank dataset \cite{buechel2022emobank}.  We used the following prompt, building from previous work \cite{amin2024wide, niu2024text}:
\\

\textit{``You are an emotionally-intelligent and empathetic agent.  You will be given a piece of text, and your task is to rate how positive the emotions expressed by the writer of the text is. Specifically, reply a number of emotional valence: from 1 to 5, how positive is the emotion? 1-very negative, 3-neutral, 5-very positive.  The second is emotional arousal: From 1 to 5, how excited is the emotion? 1-very calm, 3-normal, 5-very excited. Reply with two numbers separated by a comma for each numbered input sample. Format the output as a csv. Here are a few examples:}

\textit{Input: Adrienne shook her head and made a sound of disgust. Output: 1.62, 3.5}

\textit{Input: She bit the inside of her bottom lip, looking at me like I was breaking her poor, sweet heart. Output: 2.0, 3.5}

\textit{Input: TV game shows go interactive.  Output: 3.1, 3.1}

\textit{Input: News Baby pandas! Baby pandas! Baby pandas!  Output: 4.33, 4.22}

\textit{Input: I'm very excited we'll be on the same team for a while. Output: 4.2, 4.2}

\textit{Input: I was feeling calm and private that night. Output: 3.1, 1.8}

\textit{Input: But in fact, once news of the handover vanished from the front pages the people of Hong Kong returned to their usual topics of conversation: the economy and the price of housing. Output: 2.9, 2.1}

\textit{The following are the inputs.  Each output should be a new line.  Do not number the outputs."}
\\
\\
We evaluated the performance of GPT-4o with the prompt on the Emobank test dataset, to check that the model was capturing information about dimensional emotions.  GPT-4o with the above prompt had a concordance correlation coefficient (CCC) of 0.57 and a Pearson correlation coefficient of 0.78 for valence.  For arousal, the CCC was 0.28 and the Pearson correlation coefficient was 0.44. We also manually examined a subset of the annotated outputs to ensure quality.  Similar prompts were used in \cite{romana2025switchboard} and benchmarked with human annotations.  These methods show that the utilized model can effectively annotate the analyzed dimensions, though its ability to annotate valence was stronger (as expected for text-only input \cite{wagner2023dawn, calvo2010affect}). 

\begin{table*}[]
\centering
\label{tab:binarized_results}
\caption{Percentage of samples predicted as neutral, happy, and sad (other outputs not shown in table).  † indicates a 95\% confidence intervals (CI) of 0.03, ‡ indicates a 95\% CI of 0.04, and unmarked numbers indicate a 95\% CI $\leq$ 0.02.  Low intelligibility refers to speech that had higher ratings in the intelligibility dimension (i.e., was less intelligible) while low harsh and low monopitch refer to speech with lower ratings in those dimensions.  Within- and between- dataset differences are shown, with red indicating comparatively less frequent predictions of that category for more atypical speech and green indicating more. The best performing model for the neutral speech categories is highlighted in bold; no best performing model is designated for spontaneous speech as labels were unknown.}
\begin{tabular}{lcccccccccccc}
                                                                          & \multicolumn{3}{l}{\textbf{Odyssey}}                                                                                                                                                                                                              & \multicolumn{3}{l}{\textbf{Emotion2Vec}}                                                                                                                                                                                                          & \multicolumn{3}{l}{\textbf{Speechbrain}}                                                                                                                                              & \multicolumn{3}{l}{\textbf{\begin{tabular}[c]{@{}l@{}}OpenAI\\ GPT-4o-audio-preview\end{tabular}}}                                                                                                                                                 \\
                                                                          & \multicolumn{1}{l}{\textbf{\begin{tabular}[c]{@{}l@{}}\%\\ Neut.\end{tabular}}} & \multicolumn{1}{l}{\textbf{\begin{tabular}[c]{@{}l@{}}\%\\ Sad\end{tabular}}} & \multicolumn{1}{l}{\textbf{\begin{tabular}[c]{@{}l@{}}\%\\ Happy\end{tabular}}} & \multicolumn{1}{l}{\textbf{\begin{tabular}[c]{@{}l@{}}\%\\ Neut.\end{tabular}}} & \multicolumn{1}{l}{\textbf{\begin{tabular}[c]{@{}l@{}}\%\\ Sad\end{tabular}}} & \multicolumn{1}{l}{\textbf{\begin{tabular}[c]{@{}l@{}}\%\\ Happy\end{tabular}}} & \textbf{\begin{tabular}[c]{@{}c@{}}\%\\ Neut.\end{tabular}} & \textbf{\begin{tabular}[c]{@{}c@{}}\%\\ Sad\end{tabular}} & \textbf{\begin{tabular}[c]{@{}c@{}}\%\\ Happy\end{tabular}} & \multicolumn{1}{l}{\textbf{\begin{tabular}[c]{@{}l@{}}\%\\ Neut.\end{tabular}}} & \multicolumn{1}{l}{\textbf{\begin{tabular}[c]{@{}l@{}}\%\\ Sad\end{tabular}}} & \multicolumn{1}{l}{\textbf{\begin{tabular}[c]{@{}l@{}}\%\\ Happy\end{tabular}}} \\ \hline
\textit{\begin{tabular}[c]{@{}l@{}}Spontaneous\\ Speech\end{tabular}}     & \multicolumn{1}{l}{}                                                            & \multicolumn{1}{l}{}                                                          & \multicolumn{1}{l}{}                                                            & \multicolumn{1}{l}{}                                                            & \multicolumn{1}{l}{}                                                          & \multicolumn{1}{l}{}                                                            & \multicolumn{1}{l}{}                                        & \multicolumn{1}{l}{}                                      & \multicolumn{1}{l}{}                                        & \multicolumn{1}{l}{}                                                            & \multicolumn{1}{l}{}                                                          & \multicolumn{1}{l}{}                                                            \\
\textbf{SAP less intell.}                                                 & 0.09                                                                            & 0.73                                                                          & 0.10                                                                            & 0.26                                                                            & 0.41†                                                                         & 0.15                                                                            & 0.23                                                        & 0.10                                                      & 0.58†                                                       & 0.65†                                                                           & 0.12                                                                          & 0.20                                                                            \\
\textbf{SAP more intell.}                                                 & 0.21                                                                            & 0.56                                                                          & 0.14                                                                            & 0.43                                                                            & 0.23                                                                          & 0.25                                                                            & 0.41                                                        & 0.10                                                      & 0.29                                                        & 0.75                                                                            & 0.06                                                                          & 0.18                                                                            \\
\textit{\textbf{Diff. (more - less atypical)}}                            & {\color[HTML]{FE0000} \textit{-0.12}}                                           & {\color[HTML]{009901} \textit{0.17}}                                          & {\color[HTML]{FE0000} \textit{-0.04}}                                           & {\color[HTML]{FE0000} \textit{-0.17}}                                           & {\color[HTML]{009901} \textit{0.18}}                                          & {\color[HTML]{FE0000} \textit{-0.10}}                                           & {\color[HTML]{FE0000} \textit{-0.18}}                       & \textit{0.0}                                              & {\color[HTML]{009901} \textit{0.29}}                        & {\color[HTML]{FE0000} \textit{-0.10}}                                           & {\color[HTML]{009901} \textit{0.06}}                                          & {\color[HTML]{009901} \textit{0.02}}                                            \\
\textbf{SAP more harsh}                                                   & 0.08                                                                            & 0.75†                                                                         & 0.09                                                                            & 0.23†                                                                           & 0.38†                                                                         & 0.20†                                                                           & 0.23†                                                       & 0.10                                                      & 0.53†                                                       & 0.67†                                                                           & 0.12                                                                          & 0.17                                                                            \\
\textbf{SAP less harsh}                                                   & 0.19                                                                            & 0.58                                                                          & 0.14                                                                            & 0.41                                                                            & 0.27                                                                          & 0.22                                                                            & 0.38                                                        & 0.10                                                      & 0.35                                                        & 0.72                                                                            & 0.07                                                                          & 0.19                                                                            \\
\textit{\textbf{Diff. (more - less atypical)}}                            & {\color[HTML]{FE0000} \textit{-0.11}}                                           & {\color[HTML]{009901} \textit{0.17}}                                          & {\color[HTML]{FE0000} \textit{-0.05}}                                           & {\color[HTML]{FE0000} \textit{-0.18}}                                           & {\color[HTML]{009901} \textit{0.11}}                                          & {\color[HTML]{FE0000} \textit{-0.02}}                                           & {\color[HTML]{FE0000} \textit{-0.15}}                       & {\color[HTML]{009901} \textit{0.0}}                       & {\color[HTML]{009901} \textit{0.18}}                        & {\color[HTML]{FE0000} \textit{-0.05}}                                           & {\color[HTML]{009901} \textit{0.05}}                                          & {\color[HTML]{FE0000} \textit{-0.02}}                                           \\
\textbf{SAP more mono.}                                                   & 0.07                                                                            & 0.79                                                                          & 0.06                                                                            & 0.28†                                                                           & 0.44†                                                                         & 0.13                                                                            & 0.25                                                        & 0.14                                                      & 0.53†                                                       & 0.65†                                                                           & 0.14                                                                          & 0.18                                                                            \\
\textbf{SAP less mono.}                                                   & 0.21                                                                            & 0.55                                                                          & 0.16                                                                            & 0.41                                                                            & 0.23                                                                          & 0.25                                                                            & 0.39                                                        & 0.08                                                      & 0.33                                                        & 0.74                                                                            & 0.06                                                                          & 0.18                                                                            \\
\textit{\textbf{Diff. (more - less atypical)}}                            & {\color[HTML]{FE0000} \textit{-0.14}}                                           & {\color[HTML]{009901} \textit{0.24}}                                          & {\color[HTML]{FE0000} \textit{-0.10}}                                           & {\color[HTML]{FE0000} \textit{-0.13}}                                           & {\color[HTML]{009901} \textit{0.21}}                                          & {\color[HTML]{FE0000} \textit{-0.12}}                                           & {\color[HTML]{FE0000} \textit{-0.14}}                       & {\color[HTML]{009901} \textit{0.06}}                      & {\color[HTML]{009901} \textit{0.20}}                        & {\color[HTML]{FE0000} \textit{-0.09}}                                           & {\color[HTML]{009901} \textit{0.08}}                                          & \textit{0.00}                                                                   \\
\textbf{Switchboard}                                                      & 0.17                                                                            & 0.40                                                                          & 0.21                                                                            & 0.58                                                                            & 0.09                                                                          & 0.16                                                                            & 0.82                                                        & 0.0                                                       & 0.16                                                        & 0.79                                                                            & 0.07                                                                          & 0.07                                                                            \\
\textit{\textbf{Diff (SAP avg. - Swb.)}}                                  & {\color[HTML]{FE0000} \textit{-0.03}}                                           & {\color[HTML]{009901} \textit{0.26}}                                          & {\color[HTML]{FE0000} \textit{-0.10}}                                           & {\color[HTML]{FE0000} \textit{-0.24}}                                           & {\color[HTML]{009901} \textit{0.24}}                                          & {\color[HTML]{009901} \textit{0.04}}                                            & {\color[HTML]{FE0000} \textit{-0.51}}                       & {\color[HTML]{009901} \textit{0.10}}                      & {\color[HTML]{009901} \textit{0.28}}                        & {\color[HTML]{FE0000} \textit{-0.09}}                                           & {\color[HTML]{009901} \textit{0.03}}                                          & {\color[HTML]{009901} \textit{0.11}}                                            \\ \hline
\textit{\begin{tabular}[c]{@{}l@{}}Read\\ Sentences\end{tabular}}         &                                                                                 &                                                                               &                                                                                 &                                                                                 &                                                                               &                                                                                 & \multicolumn{1}{l}{}                                        & \multicolumn{1}{l}{}                                      & \multicolumn{1}{l}{}                                        & \multicolumn{1}{l}{}                                                            & \multicolumn{1}{l}{}                                                          & \multicolumn{1}{l}{}                                                            \\
\textbf{SAP less intell.}                                                 & 0.05                                                                            & 0.82                                                                          & 0.02                                                                            & 0.17                                                                            & 0.51†                                                                         & 0.06                                                                            & 0.17                                                        & 0.51†                                                     & 0.06                                                        & \textbf{0.43†}                                                                  & 0.17                                                                          & 0.08                                                                            \\
\textbf{SAP more intell.}                                                 & 0.09                                                                            & 0.74                                                                          & 0.01                                                                            & 0.30                                                                            & 0.40                                                                          & 0.07                                                                            & 0.30                                                        & 0.40                                                      & 0.07                                                        & \textbf{0.53}                                                                   & 0.03                                                                          & 0.14                                                                            \\
\textit{\textbf{Diff. (more - less atypical)}}                            & {\color[HTML]{FE0000} \textit{-0.04}}                                           & {\color[HTML]{009901} \textit{0.08}}                                          & {\color[HTML]{009901} \textit{0.01}}                                            & {\color[HTML]{FE0000} \textit{-0.13}}                                           & {\color[HTML]{009901} \textit{0.11}}                                          & {\color[HTML]{FE0000} \textit{-0.01}}                                           & {\color[HTML]{FE0000} \textit{-0.06}}                       & {\color[HTML]{009901} \textit{0.02}}                      & {\color[HTML]{009901} \textit{0.29}}                        & {\color[HTML]{FE0000} \textit{-0.10}}                                           & {\color[HTML]{009901} \textit{0.14}}                                          & {\color[HTML]{FE0000} \textit{-0.06}}                                           \\
\textbf{SAP more harsh}                                                   & 0.04                                                                            & 0.79†                                                                         & 0.01                                                                            & 0.17†                                                                           & 0.44‡                                                                         & 0.07                                                                            & 0.17†                                                       & 0.44‡                                                     & 0.07                                                        & \textbf{0.47‡}                                                                  & 0.10                                                                          & 0.13                                                                            \\
\textbf{SAP less harsh}                                                   & 0.09                                                                            & 0.75                                                                          & 0.01                                                                            & 0.29                                                                            & 0.43                                                                          & 0.07                                                                            & 0.29                                                        & 0.43                                                      & 0.07                                                        & \textbf{0.52}                                                                   & 0.06                                                                          & 0.12                                                                            \\
\textit{\textbf{Diff. (more - less atypical)}}                            & {\color[HTML]{FE0000} \textit{-0.05}}                                           & {\color[HTML]{009901} \textit{0.04}}                                          & \textit{0.00}                                                                   & {\color[HTML]{FE0000} \textit{-0.12}}                                           & {\color[HTML]{009901} \textit{0.01}}                                          & \textit{0.00}                                                                   & {\color[HTML]{FE0000} \textit{-0.16}}                       & {\color[HTML]{FE0000} \textit{-0.01}}                     & {\color[HTML]{009901} \textit{0.13}}                        & {\color[HTML]{FE0000} \textit{-0.05}}                                           & {\color[HTML]{009901} \textit{0.04}}                                          & {\color[HTML]{009901} \textit{0.01}}                                            \\
\textbf{SAP more mono.}                                                   & 0.02                                                                            & 0.86                                                                          & 0.01                                                                            & 0.19                                                                            & 0.55†                                                                         & 0.04                                                                            & 0.19                                                        & 0.55†                                                     & 0.04                                                        & \textbf{0.47†}                                                                  & 0.13                                                                          & 0.10                                                                            \\
\textbf{SAP less mono.}                                                   & 0.10                                                                            & 0.72                                                                          & 0.01                                                                            & 0.29                                                                            & 0.39                                                                          & 0.08                                                                            & 0.29                                                        & 0.39                                                      & 0.08                                                        & \textbf{0.52}                                                                   & 0.04                                                                          & 0.13                                                                            \\
\textit{\textbf{Diff. (more - less atypical)}}                            & {\color[HTML]{FE0000} \textit{-0.08}}                                           & {\color[HTML]{009901} \textit{0.14}}                                          & \textit{0.00}                                                                   & {\color[HTML]{FE0000} \textit{-0.10}}                                           & {\color[HTML]{009901} \textit{0.16}}                                          & {\color[HTML]{FE0000} \textit{-0.04}}                                           & {\color[HTML]{009901} \textit{0.01}}                        & {\color[HTML]{009901} \textit{0.05}}                      & {\color[HTML]{009901} \textit{0.14}}                        & {\color[HTML]{FE0000} \textit{-0.05}}                                           & {\color[HTML]{FE0000} \textit{0.09}}                                          & {\color[HTML]{FE0000} \textit{-0.03}}                                           \\
\textbf{Common Voice}                                                     & 0.90                                                                            & 0.04                                                                          & 0.01                                                                            & 0.78                                                                            & 0.04                                                                          & 0.07                                                                            & 0.78                                                        & 0.04                                                      & 0.07                                                        & \textbf{0.91}                                                                   & 0.01                                                                          & 0.04                                                                            \\
\textit{\textbf{Diff (SAP avg. - CV)}}                                    & {\color[HTML]{FE0000} \textit{-0.84}}                                           & {\color[HTML]{009901} \textit{0.74}}                                          & \textit{0.00}                                                                   & {\color[HTML]{FE0000} \textit{-0.55}}                                           & {\color[HTML]{009901} \textit{0.41}}                                          & {\color[HTML]{FE0000} \textit{-0.01}}                                           & {\color[HTML]{FE0000} \textit{-0.36}}                       & {\color[HTML]{009901} \textit{0.08}}                      & {\color[HTML]{009901} \textit{0.23}}                        & {\color[HTML]{FE0000} \textit{-0.42}}                                           & {\color[HTML]{009901} \textit{0.08}}                                          & {\color[HTML]{009901} \textit{0.08}}                                            \\ \hline
\textit{\begin{tabular}[c]{@{}l@{}}Digital Asst.\\ Commands\end{tabular}} &                                                                                 &                                                                               &                                                                                 &                                                                                 &                                                                               &                                                                                 & \multicolumn{1}{l}{}                                        & \multicolumn{1}{l}{}                                      & \multicolumn{1}{l}{}                                        & \multicolumn{1}{l}{}                                                            & \multicolumn{1}{l}{}                                                          & \multicolumn{1}{l}{}                                                            \\
\textbf{SAP less intell.}                                                 & 0.17†                                                                           & 0.69‡                                                                         & 0.03                                                                            & 0.15†                                                                           & 0.46‡                                                                         & 0.14†                                                                           & 0.15†                                                       & 0.46‡                                                     & 0.14†                                                       & \textbf{0.58‡}                                                                  & 0.23†                                                                         & 0.05                                                                            \\
\textbf{SAP more intell.}                                                 & 0.35                                                                            & 0.49                                                                          & 0.02                                                                            & 0.37                                                                            & 0.32                                                                          & 0.15                                                                            & 0.37                                                        & 0.32                                                      & 0.15                                                        & \textbf{0.83}                                                                   & 0.05                                                                          & 0.01                                                                            \\
\textit{\textbf{Diff. (more - less atypical)}}                            & {\color[HTML]{FE0000} \textit{-0.18}}                                           & {\color[HTML]{009901} \textit{0.20}}                                          & {\color[HTML]{009901} \textit{0.01}}                                            & {\color[HTML]{FE0000} \textit{-0.22}}                                           & {\color[HTML]{009901} \textit{0.14}}                                          & {\color[HTML]{FE0000} \textit{-0.01}}                                           & {\color[HTML]{FE0000} \textit{-0.10}}                       & \textit{0.01}                                             & \textit{0.34}                                               & {\color[HTML]{FE0000} \textit{-0.25}}                                           & \textit{0.18}                                                                 & \textit{0.04}                                                                   \\
\textbf{SAP more harsh}                                                   & 0.14†                                                                           & 0.70‡                                                                         & 0.02                                                                            & 0.15†                                                                           & 0.44‡                                                                         & 0.13†                                                                           & 0.15†                                                       & 0.44‡                                                     & 0.13†                                                       & \textbf{0.61‡}                                                                  & 0.17†                                                                         & 0.04                                                                            \\
\textbf{SAP less harsh}                                                   & 0.35                                                                            & 0.50                                                                          & 0.02                                                                            & 0.36                                                                            & 0.33                                                                          & 0.15                                                                            & 0.36                                                        & 0.33                                                      & 0.15                                                        & \textbf{0.81}                                                                   & 0.06                                                                          & 0.02                                                                            \\
\textit{\textbf{Diff. (more - less atypical)}}                            & {\color[HTML]{FE0000} \textit{-0.21}}                                           & {\color[HTML]{009901} \textit{0.20}}                                          & \textit{0.00}                                                                   & {\color[HTML]{FE0000} \textit{-0.21}}                                           & {\color[HTML]{009901} \textit{0.11}}                                          & {\color[HTML]{FE0000} \textit{-0.02}}                                           & {\color[HTML]{FE0000} \textit{-0.13}}                       & \textit{0.00}                                             & {\color[HTML]{009901} \textit{0.29}}                        & {\color[HTML]{FE0000} \textit{-0.20}}                                           & {\color[HTML]{009901} \textit{0.11}}                                          & {\color[HTML]{009901} \textit{0.02}}                                            \\
\textbf{SAP more mono.}                                                   & 0.11                                                                            & 0.76†                                                                         & 0.02                                                                            & 0.17†                                                                           & 0.51‡                                                                         & 0.08                                                                            & 0.17†                                                       & 0.51‡                                                     & 0.08                                                        & \textbf{0.60‡}                                                                  & 0.21†                                                                         & 0.04                                                                            \\
\textbf{SAP less mono.}                                                   & 0.36                                                                            & 0.48                                                                          & 0.02                                                                            & 0.36                                                                            & 0.31                                                                          & 0.16                                                                            & 0.36                                                        & 0.31                                                      & 0.16                                                        & \textbf{0.82}                                                                   & 0.05                                                                          & 0.02                                                                            \\
\textit{\textbf{Diff. (more - less atypical)}}                            & {\color[HTML]{FE0000} \textit{-0.25}}                                           & {\color[HTML]{009901} \textit{0.28}}                                          & \textit{0.00}                                                                   & {\color[HTML]{FE0000} \textit{-0.19}}                                           & {\color[HTML]{009901} \textit{0.20}}                                          & {\color[HTML]{FE0000} \textit{-0.08}}                                           & {\color[HTML]{FE0000} \textit{-0.12}}                       & \textit{0.00}                                             & {\color[HTML]{009901} \textit{0.38}}                        & {\color[HTML]{FE0000} \textit{-0.22}}                                           & {\color[HTML]{009901} \textit{0.16}}                                          & {\color[HTML]{009901} \textit{0.02}}                                            \\
\textbf{VCVA}                                                             & \textbf{0.90}                                                                   & 0.04                                                                          & 0.01                                                                            & 0.52                                                                            & 0.21                                                                          & 0.09                                                                            & 0.52                                                        & 0.21                                                      & 0.09                                                        & 0.88                                                                            & 0.01                                                                          & 0.01                                                                            \\
\textit{\textbf{Diff. (SAP avg. - VCVA)}}                                 & {\color[HTML]{FE0000} \textit{-0.65}}                                           & {\color[HTML]{009901} \textit{0.56}}                                          & {\color[HTML]{009901} \textit{0.01}}                                            & {\color[HTML]{FE0000} \textit{-0.26}}                                           & {\color[HTML]{009901} \textit{0.19}}                                          & {\color[HTML]{009901} \textit{0.05}}                                            & {\color[HTML]{FE0000} \textit{-0.21}}                       & {\color[HTML]{009901} \textit{0.01}}                      & {\color[HTML]{009901} \textit{0.38}}                        & {\color[HTML]{FE0000} \textit{-0.17}}                                           & {\color[HTML]{009901} \textit{0.12}}                                          & {\color[HTML]{009901} \textit{0.02}}                                           
\end{tabular}
\end{table*}
Note that the predicted ratings used for these psuedo-labels were generated by text-only input to the model while the evaluation results for GPT-4o-audio-preview included audio input to the model.  Text-only inputs were used for dimensional pseudo-labels in order to assess performance differences of the dimensional speech models in a setting where the labels were not directly impacted by the acoustic properties of the speech samples. We tabulated Pearson correlation between the predicted speech valence ratings and text valence ratings and the predicted speech arousal ratings and text arousal ratings, respectively.  We used the binarized labels to ensure sufficient data for correlation analysis.  Higher correlations signify better performance, as they correspond to more agreement between the text-only pseudo-label and the speech-only prediction.

\subsection{Strategies for improving performance}

We conducted initial investigations on improving performance of affect models on atypical speech using (1) fine-tuning, and (2) personalization via prompting. We used the GPT-4o dimensional valence scores as labels to fine-tune the Odyssey valence speech model on data from the SAP dataset.  We selected the dimensional valence model for this experiment because the GPT-4o text-only and Odyssey acoustic model had reasonable performance for valence for SAP and Switchboard.  The data marked as training in the dataset \cite{hasegawa2024community} was used for training, and the validation and test splits were used for reporting results.  We fine-tuned the speech emotion recognition layers of the model using a mean squared error (MSE) loss for 50 epochs with an Adam optimizer and a learning rate of 0.0001

For the personalization via prompting experiment, we provided GPT-4o with personalized information about the speaker's atypical speech properties in the prompt.  For each sample, we included binarized information about the annotations in the prompt as follows:  
\textit{Please consider this speaker's unique voice characteristics when assessing their speech for emotion. This speaker's voice is ['more harsh', 'more monopitch', 'less intelligible'] than typical speech.}  For each sample, we included all descriptions that applied from: 'more harsh', 'more monopitch', and 'less intelligible'.  We analyzed the impact of using personalized prompts on the read speech categories, where the speech was expected to be neutral.  We did not assess personalization with spontaneous speech because the affective labels of spontaneous speech were unknown so we would not be able to assess if performance changes were improvements or degradations.

\section{Results and Discussion}
\begin{figure}[t!]
  \centering
  \includegraphics[width=\linewidth]{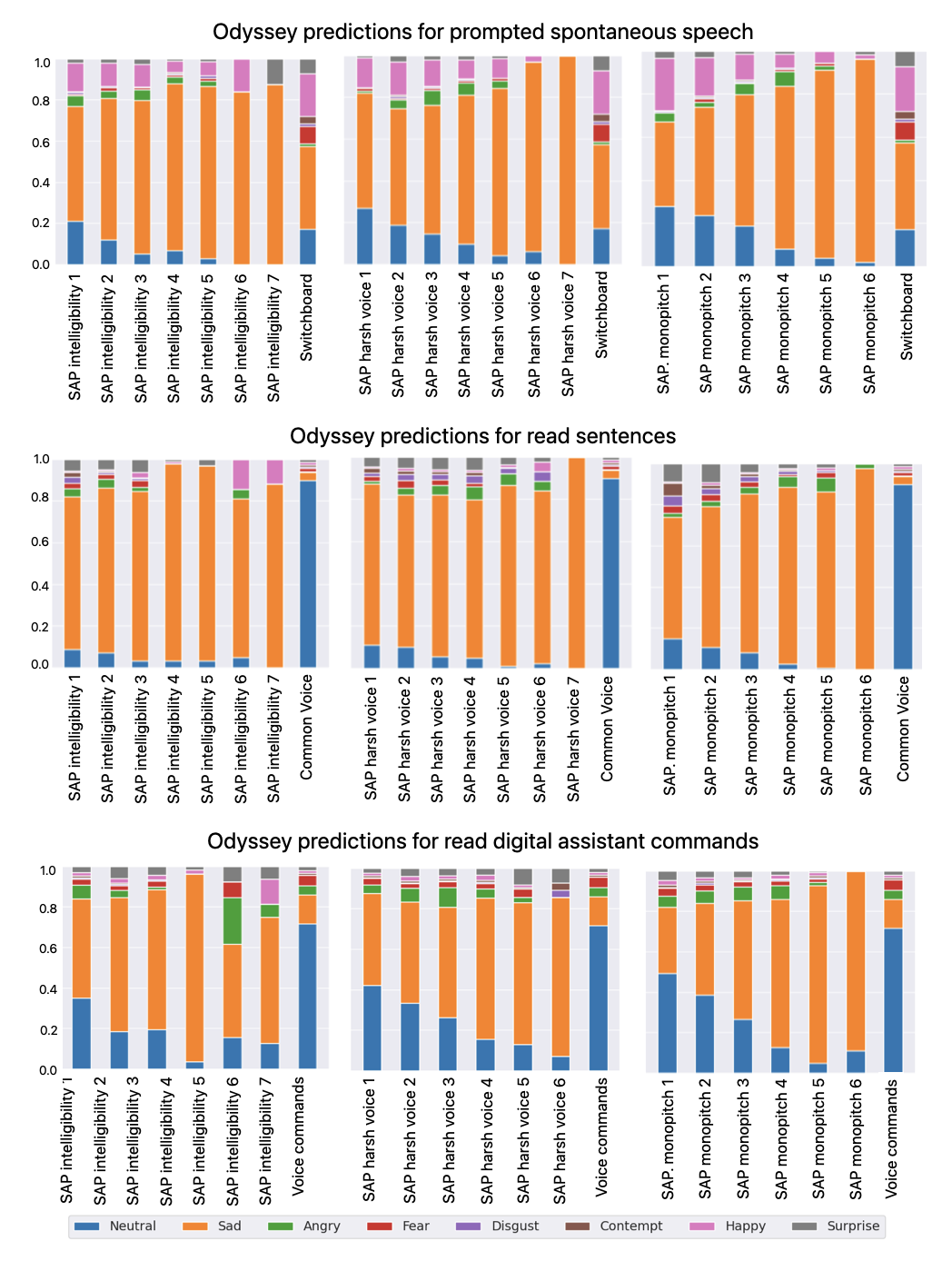}
  \caption{Distribution of each categorical emotion percentage model outputs for the Odyssey model for each speech category}
  \label{fig:odyssey_categorical_all}
\end{figure}

\subsection{Off-the-shelf Model Performance}

Table I shows the percentage of samples predicted as neutral, happy, and sad (generally the most frequently predicted emotions) for binarized subsets of the SAP dataset and typical speech datasets for the evaluated categorical emotion models, including differences in prediction frequency within- and between-datasets.  Red numbers indicate comparatively less frequent predictions of an emotion for more atypical speech and green indicates comparatively more frequent predictions of an emotion for more atypical speech. Within the SAP dataset, there were clear differences in prediction distributions between levels of atypicality.  For Odyssey and Emotion2Vec, the percentage of predicted sad speech was significantly higher for speech that was less intelligible, more harsh, and had more monopitch.  GPT-4o-audio-preview also often predicted higher proportions of sad speech for speech that was more atypical, though the differences were less stark than for Odyssey and Emotion2Vec.  The Speechbrain model predicted a higher percentage of happiness and a lower percentage of sadness for more atypical speech data.  The differences in which categorical emotion was most impacted by atypical speech between the models could be due to differences in training data -- Speechbrain was the only acoustic model trained only on acted affect data.

For the neutral read speech samples, all models predicted neutral less frequently for samples from SAP than for samples from typical speech datasets, suggesting that models confused atypical speech properties with affect.  The difference was most severe for the Odyssey model, followed by Emotion2Vec and Speechbrain.  GPT-4o-audio-preview predicted neutral significantly more frequently than the other models, suggesting more robustness to atypical speech.  However, GPT-4o was sometimes confused by the content of the sample and was unable to make predictions for a small number of samples and would instead generate outputs like ``\textit{I'm sorry, I can't adjust the volume."}. As part of generating annotations, GPT-4o-audio-preview has transcription capabilities and may have used lexical information in making predictions, which could have helped improve robustness for samples with atypical speech.

\begin{figure}[t!]
  \centering
  \includegraphics[width=\linewidth]{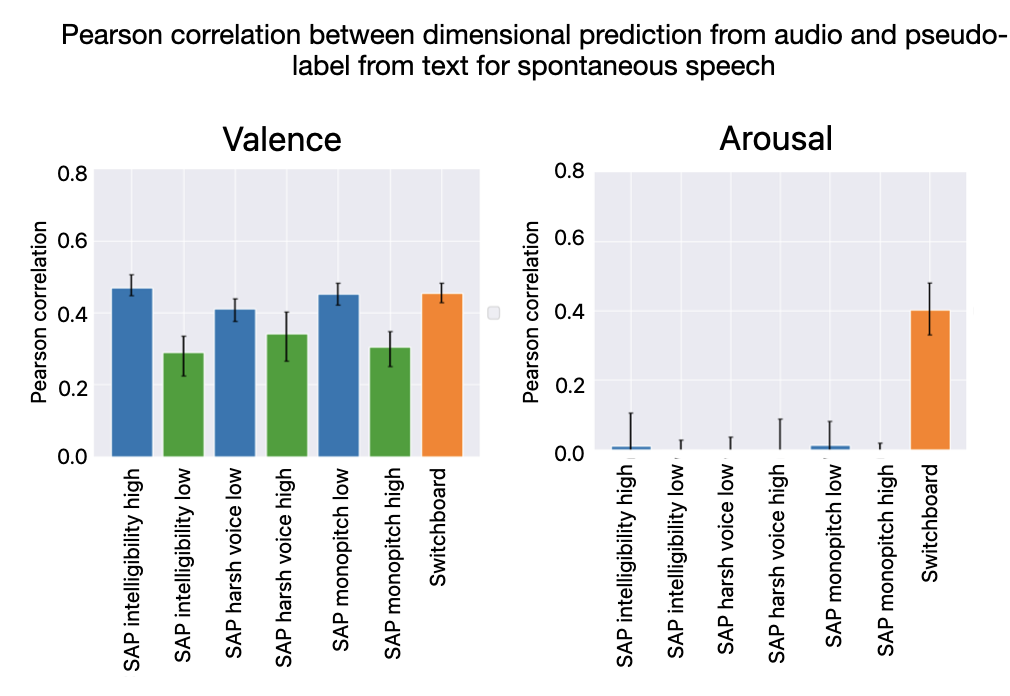}
  \caption{Correlations for Odyssey dimensional predictions and pseudo-labels from GPT-4o with 95\% confidence intervals calculated via bootstrapping.  SAP results reported across all splits.}
  \label{fig:dimensional_correlations}
\end{figure}

We also looked at trends between non-binarized ratings.  Figure \ref{fig:odyssey_categorical_all} shows results for the Odyssey categorical model.  Similar types of patterns relative to the presented binarized results were observed for all models.   

Within the SAP dataset, samples with higher rated atypicalities had a smaller percentage of samples rated as neutral for all speech categories and rating types.  Trends were generally similar among types of atypicalities with some nuances -- for instance, monopitch was particularly tied to less happiness for spontaneous speech data.  The read speech categories (digital assistant commands and sentences) had considerably lower percentages of samples rated as neutral in the SAP dataset at every rating, even at very low ratings of atypicality.  The low percentage of SAP samples rated as neutral suggests that even  mild atypicalities were confused with affect by acoustic models.  For prompted spontaneous speech, the percentage of neutral speech was similar for the Switchboard data and for the SAP data that had only mild ratings - suggesting some similarities in distributions of affective content among the datasets.  Samples with any atypical intelligibility (rating $>=$2) had considerably low percentages of neutral ratings compared to the Switchboard data.  Atypical harshness and atypical pitch variations also corresponded to a reduced percentage of predicted neutral speech, but less sharply than atypical intelligibility.  

\begin{table}[]
\centering
\label{tab:finetuning}
\caption{Pearson correlations with 95\% CIs (calculated via bootstrapping) for predicted valence before and after fine-tuning, using GPT-4o pseudo-labels.  SAP performance is reported on validation and test splits}
\begin{tabular}{lccl}
                                                                              & \multicolumn{2}{c}{Pearson correlation}                                                                                           &                                                       \\
\multicolumn{1}{l|}{}                                                         & \begin{tabular}[c]{@{}c@{}}Odyssey\\ valence\end{tabular}   & \begin{tabular}[c]{@{}c@{}}Odyss. valence,\\ FT on SAP\end{tabular} & \begin{tabular}[c]{@{}l@{}}Perf.\\ Diff.\end{tabular} \\ \hline
\multicolumn{1}{l|}{\begin{tabular}[c]{@{}l@{}}Intellig.\\ high\end{tabular}} & \begin{tabular}[c]{@{}c@{}}0.51\\ (0.46, 0.56)\end{tabular} & \begin{tabular}[c]{@{}c@{}}0.56\\ (0.63, 0.61)\end{tabular}         & +0.06                                                 \\
\multicolumn{1}{l|}{\begin{tabular}[c]{@{}l@{}}Intellig.\\ low\end{tabular}}  & \begin{tabular}[c]{@{}c@{}}0.22\\ (0.08, 0.14)\end{tabular} & \begin{tabular}[c]{@{}c@{}}0.31\\ (0.23, 0.39)\end{tabular}         & +0.10                                                 \\
\multicolumn{1}{l|}{\begin{tabular}[c]{@{}l@{}}Harsh\\ low\end{tabular}}      & \begin{tabular}[c]{@{}c@{}}0.37\\ (0.29, 0.42)\end{tabular} & \begin{tabular}[c]{@{}c@{}}0.45\\ (0.39, 0.51)\end{tabular}         & +0.08                                                 \\
\multicolumn{1}{l|}{\begin{tabular}[c]{@{}l@{}}Harsh\\ high\end{tabular}}     & \begin{tabular}[c]{@{}c@{}}0.36\\ (0.29, 0.46)\end{tabular} & \begin{tabular}[c]{@{}c@{}}0.45\\ (0.38, 0.54)\end{tabular}         & +0.08                                                 \\
\multicolumn{1}{l|}{\begin{tabular}[c]{@{}l@{}}Monopitch\\ low\end{tabular}}  & \begin{tabular}[c]{@{}c@{}}0.43\\ (0.37, 0.49)\end{tabular} & \begin{tabular}[c]{@{}c@{}}0.50\\ (0.45, 0.56)\end{tabular}         & +0.07                                                 \\
\multicolumn{1}{l|}{\begin{tabular}[c]{@{}l@{}}Monopitch\\ high\end{tabular}}  & \begin{tabular}[c]{@{}c@{}}0.30\\ (0.22, 0.38)\end{tabular} & \begin{tabular}[c]{@{}c@{}}0.38\\ (0.31, 0.46)\end{tabular}         & +0.08                                                 \\
\multicolumn{1}{l|}{\begin{tabular}[c]{@{}l@{}}Switch\\ board\end{tabular}}   & \begin{tabular}[c]{@{}c@{}}0.45\\ (0.43, 0.48)\end{tabular} & \begin{tabular}[c]{@{}c@{}}0.46\\ (0.43, 0.50)\end{tabular}         & +0.01                                                
\end{tabular}
\end{table}

Atypical speech also influenced the correlation strength between dimensional emotion predictions and GPT-4o pseudo-labels (Figure \ref{fig:dimensional_correlations}).  For each type of atypicality for valence, the correlations for data with low atypicality were similar to the correlations for the Switchboard data.  The valence correlations for the speech with high atypicalities were lower than the Switchboard data, significantly so for low intelligibility speech and speech with high monopitch.  The arousal correlations for all SAP data were significantly lower than the correlation for the Switchboard data, regardless of the amount of atypicality.  The low arousal correlations in SAP may be influenced by both speech atypicalities and dataset content.  It is possible that the samples in the SAP data may have been inherently more difficult to rate with respect to arousal than the Switchboard data due to differences in content.

\subsection{Improving performance for atypical speech}

Table II compares the performance of the Odyssey dimensional valence model before and after fine-tuning on the SAP training data using the GPT-4o valence pseudo-labels.  Fine-tuning consistently improved performance for all types of atypical speech.  Additionally, fine-tuning did not negatively impact performance on Switchboard, the typical speech dataset.  The clear performance improvements with a simple fine-tuning strategy and a small pseudo-labeled dataset highlights the potential for improving model robustness for atypical speech by collecting and curating diverse datasets.  A larger dataset of affectively-elicited speech from people with speech atypicalities along with careful labels could likely significantly model improve performance for atypical speakers, and may also improve model performance generally.

Personalized prompting led to only minor changes in distributional predictions for the read speech categories: the percent of neutral speech increased from 51 ± 2\% to 53 ± 1\% for read digital assistant commands and from 78 ± 1\% to 81 ± 1\% for read digital assistant commands.  The slight increase in predicted neutral speech suggests minor improvement on robustness by personalized prompting, though future work is needed to expand on assess the significance of these results.   

\section{Limitations and Future Work}

Our analysis was limited to English datasets, and additional work is also needed to characterize performance differences with atypical speech in other languages.   We attempted to match the domain of atypical speech datasets and typical speech datasets to allow for comparisons, but there were still some data distributional differences between datasets that could have impacted comparisons.  Given the available knowledge about dataset content and the within-dataset comparisons, the reported results still capture real trends related to generalization.  While the SAP data included speakers with a number of different etiologies, it is still a non-comprehensive dataset of additional speech.  Including broader speaker diversity (e.g., people with structural voice conditions, aging populations) would expand understanding of affect and voice properties.  In the future, collecting a dataset of affective speech from both typical and atypical speakers of diverse demographics and etiologies would allow for additional comparative analyses and enable improved training strategies.

We used pseudo-labels from GPT-4o for the dimensional performance analysis. While previous work \cite{niu2024rethinking} and our Emobank evaluation validates the use of these labels, they may have limitations and different biases than annotator-generated data or self-labeled data.  Analyses on differences between self-labeled affective emotional speech data and annotator-labeled data will be valuable in understanding how to build robust models and in assessing human errors from listeners interpreting affect in atypical speech.  Since GPT-4o did not have access to speech information, its performance in generating pseudo-labels was not impacted by how each sample was spoken.  However, lexical differences correlating to the atypical speech rating (e.g., utterance length differences) could have influenced the GPT-4o annotations, and were not investigated in this work.  Still, from our analyses of word counts, the distributions of word counts were generally similar between samples of different ratings.  Additionally, GPT-4o-audio-preview may use lexical content in generating predictions, which could be related to the low performance change when using personalized prompting.  The extent and impact of this can be assessed in future work.

\section{Conclusions}
\textbf{Weak generalization} Pre-trained affect models consistently had weak generalizability to atypical speech data across both dimensional and categorical emotions.  The distribution of categorical ratings for atypical speech data was significantly different between levels of atypicality and when compared to typical speech data.  Similar trends were observed for each type of analyzed atypicality, with some differences in the type and magnitude of impact (e.g., a clearer relationship between happiness and monopitch for spontaneous speech than with happiness and other dimensions). Neutral read speech from speakers with speech atypicalities was consistently predicted as sad while neutral samples from datasets with typical speech were consistently predicted correctly as neutral.  Correlations for dimensional predictions with pseudo-labels were also lower for atypical speech than atypical speech, particularly for low intelligibility vs. high intelligibility speech.

\textbf{Data diversity} The weak generalization to atypical speech may be related to lack of representation of atypical speech in affective datasets.  The analyzed models were trained on acted emotional dataset and podcast datasets that did not specifically include atypical speech.  The lack of exposure to this type of data during training combined with the added difficulty of differentiating affective variations from speaker atypicalities likely contributes to the observed performance gap.  Speechbrain, a model trained directly only on acted speech, predicted 'happy' more frequently for more atypical speech while other models trained on more naturalistic datasets predicted 'sad' more frequently for atypical speech - this suggests a potentially strong impact on the affect elicitation strategy of training data on the relationship between learned acoustic cues and affect categories. The results highlight the need for collecting a unified dataset of diverse samples of affective speech from both typical and atypical speakers with self- and annotator-labeled affect.

\textbf{Limitations of annotated labels} Additionally, models trained on interpreted affect labels may have inherent robustness limitations due to overlap in voice properties associated with affect expression by actors, voice properties impacting affect perception by raters, and voice properties modulated by atypical speech and speaker properties.  There are likely organic variations in these dimensions (e.g., harshnesss, pitch) tied to both atypical speech and affect, as well as labeling bias tied to voice and speech properties.  Datasets used to train models are generally labeled by external annotators who only have access to a speech sample, often from an acted- or domain-specific dataset.  Because contextual information about the situation and speaker is not present, labelers may overfit to acoustic variations in the dataset that might actually be due to other, non-emotional factors. Models that directly track changes in perceptual descriptions of voice properties \cite{interspeechanonymous} - without using interpreted affect labels that may not account for key speaker and contextual properties - may be strong candidates for interpretable, interactive systems.

\section{Ethical Impact Statement}
We evaluated the robustness of affect models on three dimensions of atypical speech (intelligibility, harshness, and monopitch) using English data.  The presented conclusions may not generalize to non-English datasets, where atypical speech properties and affective expression may intersect with language differently.  We did not investigate intersectionality between atypical speech and other speaker characteristics (e.g., gender, accents, demographics) - an important consideration for future work.  Additionally, while the SAP dataset includes speakers with a number of etiologies, the data does not include all types of atypical speech.  Analyses including additional speaker etiologies, demographics, languages, and intersections thereof would further expand understanding of fairness and affect modeling.

Our investigation used the SAP dataset for atypical speech, which has some spontaneous speech with unlabeled affective variations.  The free-form nature of this dataset without conventional labels may limit the generalization of the results when compared to traditionally collected datasets.  Still, it is important to note that these traditionally collected and labeled datasets have limitations \cite{hernandez2021guidelines} that may be particularly prohibitive for atypical speech.  In affect-focused work with atypical speech, it is particularly important to collect self-labeled data or contextually labeled data because external annotators labeling audio-only data may not be able to correctly isolate affective variations from speaker-specific properties \cite{johnson2023recanvo}.  There is a need for broad, inclusive data collection efforts as well as personalized modeling approaches to ensure affectively-aware systems are fair for all users. 

\bibliographystyle{IEEEtran}
\bibliography{mybib}

\end{document}